\documentclass{article}
\usepackage{amsmath,graphicx,mlspconf}
\usepackage{xcolor,url,booktabs}
 \usepackage{enumitem}

\copyrightnotice{979-8-3503-2411-2/25/\$31.00 {\copyright}2025 European Union}

\toappear{2025 IEEE International Workshop on Machine Learning for Signal Processing, Aug.\ 31-- Sep.\ 3, 2025, Istanbul, Turkey}

\title{Comparison of End-to-end Speech Assessment Models\\for the NOCASA 2025 Challenge}
\name{%
    Aleksei Žavoronkov
    \qquad Tanel Alumäe}%

\address{%
    Tallinn University of Technology, Estonia
}

\begin{document}

\maketitle

\begin{abstract}
This paper presents an analysis of three end-to-end models developed for the NOCASA 2025 Challenge, aimed at automatic word-level pronunciation assessment for children learning Norwegian as a second language. Our models include an encoder-decoder Siamese architecture (E2E-R), a prefix-tuned direct classification model leveraging pretrained wav2vec2.0 representations, and a novel model integrating alignment-free goodness-of-pronunciation (GOP) features computed via CTC. We introduce a weighted ordinal cross-entropy loss tailored for optimizing metrics such as unweighted average recall and mean absolute error. Among the explored methods, our GOP-CTC-based model achieved the highest performance, substantially surpassing challenge baselines and attaining top leaderboard scores. 
\end{abstract}

\section{Introduction}
\label{sec:intro}

The task of speech pronunciation assessment focuses on automatically evaluating a language learner's pronunciation of phonemes, words, or complete utterances. Such systems can be used to provide feedback in computer-aided language learning  applications.

The Non-native Children’s Automatic Speech Assessment (NOCASA) Challenge \cite{getman2025non} was designed to benchmark the current state-of-the-art in automatic word-level pronunciation assessment for children learning Norwegian as a second language (L2).
The organizers released a training corpus consisting of 44 speakers and 10,334 utterances. Each utterance contains a recording of one of 205 target words. The speakers were children aged 5–12, including native (L1) speakers, beginner L2 learners of Norwegian, and children with no prior exposure to the language. For each utterance in the training set, a pronunciation accuracy score ranging from 1 to 5 assigned by a human expert is provided, along with the orthographic transcription of the prompted word. The test set contains 1,930 utterances from 8 speakers. Word transcripts are provided for the test set as well.

The goal for challenge participants was to develop an automatic speech assessment system capable of predicting the pronunciation score for each utterance in the test set. Participants could submit their predictions to an evaluation server, with a maximum of five submission attempts allowed.

This paper describes the models developed by the TalTech team for the NOCASA Challenge. We investigated three different end-to-end architectures for word-level speech assessment. Our best-performing model utilized interpretable, character-level goodness-of-pronunciation (GOP) features, derived from CTC emission probabilities obtained using a pretrained speech recognition model. This approach builds upon the method introduced in \cite{cao24b_interspeech}, but extends it by employing a fully end-to-end trainable framework and incorporating additional character embedding features.

We also experimented with different loss functions and propose the use of a weighted ordinal cross-entropy (CE) loss that balances performance across both unweighted average recall (UAR) and mean absolute error (MAE). Our best models clearly outperformed the baselines provided by the challenge organizers and achieved the best scores in each metric among all participants.

\section{Methods}
\label{sec:format}

\subsection{Models}

\subsubsection{Encoder-decoder Siamese model (E2E-R)}

\begin{figure}
    \centering
    \includegraphics[width=1.0\columnwidth]{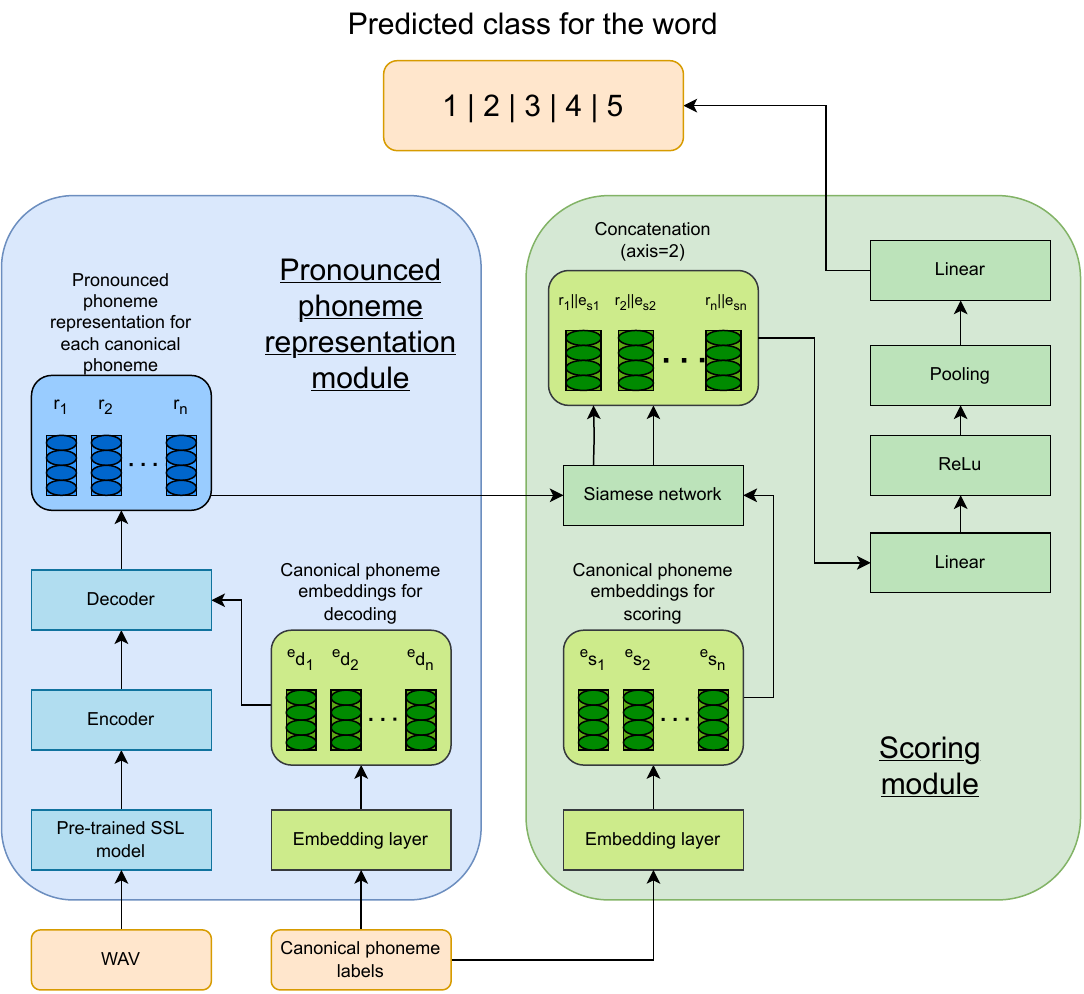}
    \caption{Architecture of the encoder-decoder Siamese model (E2E-R)}
    \label{fig:e2er-arch}
\end{figure}

We adapted the E2E-R model proposed in \cite{10255657}, which follows a two-stage training strategy (see Figure \ref{fig:e2er-arch}). In the first stage, a pre-trained self-supervised learning (SSL) model is fine-tuned for phoneme recognition using a hybrid encoder-decoder CTC-attention mechanism. The second stage introduces a Siamese neural network that compares embeddings of pronounced phonemes with their canonical counterparts to compute pronunciation scores. This scoring module is trained using phoneme-level annotated utterances. The architecture eliminates the need for complex feature engineering and external forced-alignment tools. On a standard English dataset, the model demonstrates performance comparable to state-of-the-art systems.

The original model was designed to predict a score for each pronounced phoneme. However, for the challenge, we require the model to produce a single score for the entire utterance. To address this, we modified the architecture of the model’s scoring component.

The phoneme recognition module remains unchanged from the original design. However, a small modification was made to the Siamese network within the scoring module: Layer Normalization was used instead of Batch Normalization due to training instability.

In the original model, scoring was based on a similarity comparison between two vectors --- one representing the predicted phoneme and the other representing the canonical phoneme. A similarity function was used to produce scores for each phoneme.
In our version, this similarity function is replaced with a different approach. The phoneme embeddings are first concatenated and passed through a linear layer with a ReLU activation. The output is then processed using max pooling over the actual sequence lengths. Finally, the result of max pooling is passed through another linear layer, which produces the class probabilities.

We used model provided by Nasjonalbiblioteket AI Lab (NbAiLab) as our SSL model\footnote{\url{https://huggingface.co/NbAiLab/nb-wav2vec2-300m-bokmaal}}. We used version with 300 millions parameters. We fine-tuned the model for the letter recognition, as opposed to phoneme recognition, as in the original E2E-R model. 
This decision was made because provided data were annotated only on the letters level.

For the finetuning we used NB Tale dataset \cite{nbtale2013}. NB Tale is a Norwegian acoustic-phonetic speech database. It consists of 3 modules: Module 1 - manuscript-read speech L1, Module 2 - manuscript-read speech L2, Module 3 - spontaneous speech. For our experiments, we utilized modules 1 and 2 of the dataset. Both are annotated (time-stamped) at a phonotypic level of detail. The dataset is divided into two independent parts: 1) training set and 2) test set. Each speaker has read 20 sentences. These are divided into three groups: A, B and C. 3 sentences come from set A (calibration set) and are read by all speakers. 12 sentences come from set B and are read by three different speakers. 5 sentences come from set C and are unique for each speaker. From the original dataset we have used 7392 utterances, 5283 for the training set and 2109 for the test set. The average utterance length is 9 seconds.

\subsubsection{Prefix-tuned direct classification model}

\begin{figure}
    \centering
    \includegraphics[width=0.5\columnwidth]{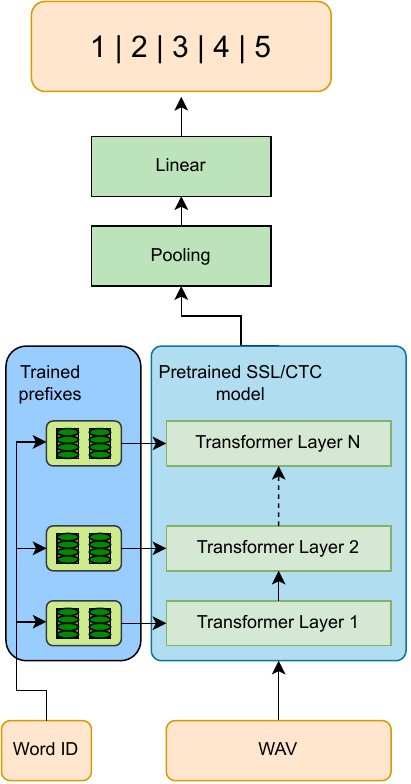}
    \caption{Architecture of the prefix-tuned audio classification model.}
    \label{fig:prefix-tuning-arch}
\end{figure}

The second model follows a standard architecture commonly used for fine-tuning a pretrained wav2vec2.0 model for audio classification. Frame-level representations from the wav2vec2.0 encoder \footnote{\url{https://huggingface.co/NbAiLab/nb-wav2vec2-1b-bokmaal}} are passed through an attentive statistics pooling layer, followed by a linear layer that outputs class logits. However, this architecture does not take into account the specific word the speaker was prompted to pronounce.

To incorporate this word information, we use prefix-tuning \cite{li2021prefixtuning}, as illustrated in Figure~\ref{fig:prefix-tuning-arch}. In this approach, each word type in the training data is associated with a list of vectors that serve as fixed, word-specific inputs to the model alongside the audio features. Specifically, we inject $L_{\text{prefix}} = 2$ prefix vectors into the model, parameterized by $\theta_p$. Following \cite{li2021prefixtuning}, the prefix vectors are prepended to each Transformer layer as additional key and value vectors. These vectors do not use positional encoding, and their corresponding outputs are discarded after the Transformer layer. The actual letter sequence of a word is not used in this model -- each word type is simply treated as a separate category.

The dimensionality of $\theta_p$ is $L_{\text{prefix}} \times \text{\#Layers} \times \text{hidden dim} \times 2$. For example, when using the large XLS-R 1B wav2vec2.0 model with a hidden size of 1280 and 48 Transformer layers, and setting the prefix length to 2, this results in approximately 122K additional parameters per word type. A related method was used in \cite{alumae2023dialect} to incorporate dialect information into a wav2vec2.0-based speech recognition model. Unlike that work, where the prefix vectors were adapted to a pretrained model, we train the prefix vectors jointly with the rest of the model.

\subsubsection{Alignment-free CTC feature based model  (GOP-CTC-AF-E2E)}

\begin{figure}
    \centering
    \includegraphics[width=1\columnwidth]{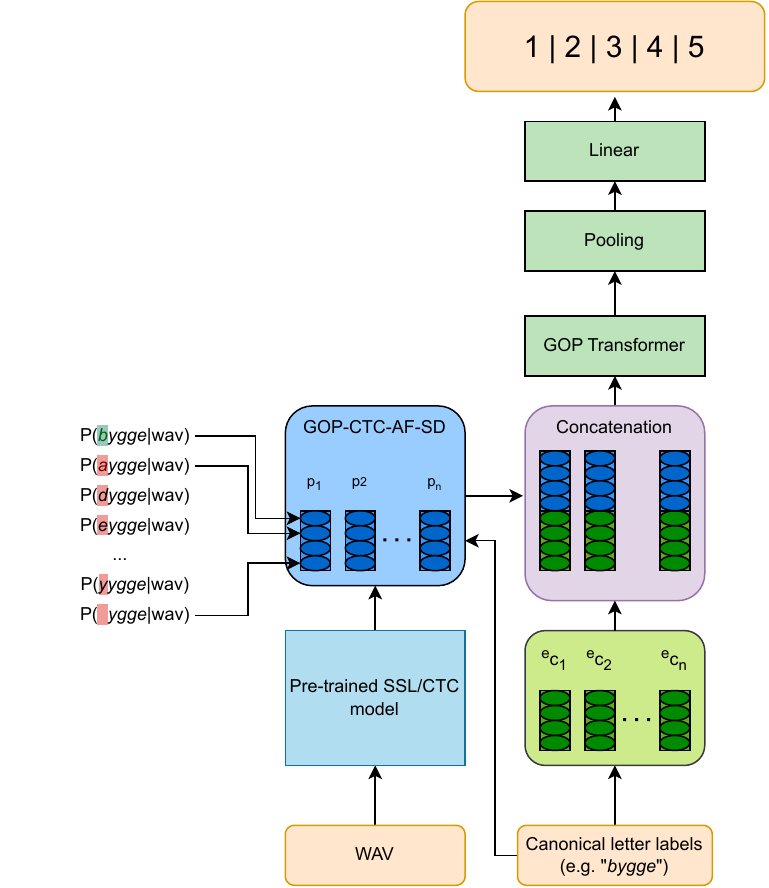}
    \caption{Architecture of the end-to-end model using CTC-based alignment-free goodness-of-pronunciation features  (GOP-CTC-AF-E2E).}
    \label{fig:ctc-gop-arch}
\end{figure}

The third model (GOP-CTC-AF-E2E) is inspired by the  alignment-free CTC GOP feature extractor proposed in \cite{cao24b_interspeech}. The features aim to identify, whether the speaker replaced some phoneme in the word's canonical pronunciation with an alternative one, or did not pronounce some phoneme at all.  In the current work, we use the similar idea in an end-to-end model which allows to finetune the underlying pretrained CTC model for speech assessment. The model is depicted on Figure \ref{fig:ctc-gop-arch}.

The alignment-free CTC substitution-deletion (GOP-CTC-AF-SD) features are computed as follows: a pretrained  CTC-based speech recognition model\footnote{\url{https://huggingface.co/NbAiLab/nb-wav2vec2-1b-bokmaal}} using letters as basic units is first used to compute frame-level emission log probabilities $X$ for all letters (and the CTC ``blank'' symbol). Based on the emissions, the following features are computed for each letter in the canonical input sequence $L_{canonical}$ of length $S$: 
\begin{itemize}[noitemsep,topsep=2pt,parsep=5pt,partopsep=2pt]
    \item Log Posterior Probability (LPP) of the canonical sequence: The LPP of the entire canonical transcription given the audio is calculated using the CTC forward algorithm:
\[LPP = \log P_{CTC}(L_{canonical} | X)\]
This global score reflects the overall acoustic likelihood of the target pronunciation. While calculated once per utterance, its value is used as a feature component for each token.
\item Log Posterior Ratios (LPR) for substitutions: For each letter $c_i$ and every letter $v_j$ in the model's vocabulary $V$, a substituted sequence $L_{sub}(i, j)$ is formed by replacing $c_i$ with $v_j$. The LPR for substitution is:
\[LPR_{sub}(i, j) = LPP - \log P_{CTC}(L_{sub}(i, j) | X)\]
This results in a vector of LPRs for $c_i$ of dimensionality $|V|$, indicating how well $c_i$ acoustically fits compared to all possible alternatives at its position.
  \item Log Posterior Ratio (LPR) for deletion: A deleted sequence $L_{del}(i)$ is formed by removing $c_i$ from $L_{canonical}$. The LPR for deletion is:
\[
LPR_{del}(i) = LPP - \log P_{CTC}(L_{del}(i) | X).
\]
This scalar value indicates the acoustic importance of $c_i$'s presence in the sequence.
\end{itemize}
All log $P_{CTC}(L | X)$ terms are computed using the standard CTC loss function (interpreted as a negative log-likelihood) on the frame-level log-probabilities. This process is alignment-free.

Simultaneously, each canonical letter $c_i$ is passed through a learnable embedding layer that maps a letter to a vector. This provides a learned semantic representation of the target letter.

For each canonical letter $c_i$, the calculated GOP features and its embedding are concatenated to form a comprehensive feature vector $F_i$:
\[
F_i = [LPP, \vec{LPR}_{sub}(i), LPR_{del}(i), \vec{Emb}_i].
\]
This results in a sequence of vectors $F_{seq} = {F_1, F_2, ..., F_S}$.

The sequence $F_{seq}$ is processed by a dedicated Transformer encoder layer (referred to as the ``GOP Transformer'').
This layer captures contextual dependencies and interactions among the token-level pronunciation features, allowing the model to learn higher-level patterns related to pronunciation quality across the token sequence.
To obtain a fixed-size representation for the entire utterance, max pooling is applied across the sequence dimension of the GOP Transformer's output. This utterance embedding is then passed through a final linear layer followed by a softmax  function to produce the utterance classification posterior probabilities.

The model is trained end-to-end.  Gradients propagate back through the entire architecture, allowing all learnable components (token embeddings, GOP Transformer, classifier, and  the base CTC model to be updated. 

At first sight, it might seem that the model is computationally expensive, since a lot of CTC-based features have to be calculated for each letter in the canonical sequence. However, the most expensive step of computing the CTC emissions is performed only once per utterance and computing CTC marginalized likelihoods for $S \times |V|$ different alternative sequences is actually  fast. Thus, the model is not substantially slower than the second model that we proposed.

\subsection{Loss function}

The challenge used multiple metrics: unweighted average recall (UAR) as the primary metric, F1 score, accuracy and mean average error (MAE). When developing our models, we mostly tried to optimize UAR and MAE. Thus, we used weighted ordinal CE as the loss function that explicitly incorporates the ordinal structure of the target labels. This loss penalizes prediction errors proportionally to their ordinal distance from the true label, ensuring that small deviations (e.g., predicting 4 instead of 5) incur lower penalties than large ones (e.g., predicting 1 instead of 5). The loss is computed based on the negative log of the complement of predicted class probabilities, scaled by a distance-based penalty and optional class-specific weights to account for label imbalance. 

The loss for a single sample is defined as:
\[
\mathcal{L}_{\text{ordinal}}(P, y) = \sum_{i=1}^{N} w_y \cdot \left[ -\log(1 - p_i) \cdot d(y, i)^\alpha \right]
\]
where  \( P = (p_1, \dots, p_N) \) is the predicted probability distribution over \( N \) classes, \(y \in \{1, \dots, N\}\) is the true class label,
\( d(y, i) \) is the absolute distance between the true class and class \( i \),
 \( \alpha \geq 0 \) controls the penalty scaling for distant errors,
 \( w_y \) is the optional class weight for the true class \( y \).
 This loss encourages predictions that are not only correct but also close to the true class when errors occur, while optionally giving more importance to underrepresented classes. With $\alpha = 0$, this loss reduces to the simple (weighted) CE loss.

\section{Experimental results}

\begin{table*}[tb]
\centering
\label{tab:dev-results}
\caption{Results of different models on the internal development set.}
\begin{tabular}{l|c|cccc}
\toprule
              & Loss                &UAR$\uparrow$  (\%)  & F1$\uparrow$ (\%) & Accuracy$\uparrow$ (\%) & MAE$\downarrow$   \\ \midrule
E2E-R         & Cross entropy (CE)  & 40.3      & 39.3    & 47.4          & 0.700     \\
Prefix-tuning & Weighted ordinal CE & 42.8      & 42.6    & 41.7          & 0.750     \\
GOP-CTC-AF-E2E    & Weighted ordinal CE & \textbf{44.0}      & \textbf{54.6}    &\textbf{ 54.7}          & \textbf{0.525}     \\ \bottomrule
\end{tabular}
\end{table*}

\subsection{Data}

The dataset provided by the challenge organizers consists of 7857 labeled training utterances and 1460 unlabeled test utterances. Due to privacy constraints, speaker identities are not disclosed. To enable model tuning, we split the training data into internal training and development sets. To prevent overfitting, it is crucial to perform this split at the speaker level. Since speaker labels were not available, we employed a speaker recognition model to cluster utterances based on similarity, thereby generating speaker pseudo-labels.

Speaker embeddings were extracted using the Wespeaker toolkit \cite{wang2023wespeaker, wang2024advancing}, specifically the SimAM-ResNet34 model\footnote{\url{https://github.com/wenet-e2e/wespeaker/blob/master/docs/pretrained.md}} \cite{qin2022simple}, which was pre-trained on the VoxBlink dataset \cite{lin2024voxblink2100kspeakerrecognition} and further finetuned on VoxCeleb2 \cite{chung2018voxceleb2}. Before embedding extraction, all utterances were volume-normalized and processed using a voice activity detection (VAD) system based on the Silero VAD model \cite{SileroVAD} to remove excessive silence at the beginning and end of each utterance.

To cluster the utterance-level speaker embeddings, we first center the data by subtracting the global mean embedding.   A pair-wise cosine-similarity matrix \(S\) is then computed and sparsified: for each embedding we preserve only its most similar neighbours, controlled by a pruning parameter \(p=0.01\), and symmetrize the result to obtain an affinity matrix \(A\). We convert \(A\) to the unnormalised graph Laplacian \(L = D - A\), where \(D\) is the diagonal degree matrix. Spectral clustering is performed by eigen-decomposing \(L\); the number of clusters is automatically estimated via the first large ``eigengap'' within a user-defined range \([K_{\min}=40, K_{\max}=45]\). The first \(K\) eigenvectors form a low-dimensional spectral embedding in which points belonging to the same speaker lie close to one another. Finally, \(K\)-means partitions this spectral space to yield the speaker labels. The estimated range of the number of speakers in the training set was calculated based on the information about the dataset published in \cite{olstad-etal-2024-collecting}.

The utterance clustering process resulted in 40 pseudo-speaker clusters. We randomly selected 20\% of the pseudo-speakers, corresponding to 1462 utterances, for the internal development set. The remaining 6395 utterances were used for training.

\subsection{Training details}

\begin{table*}[]
\centering
\label{tab:eval-results}
\caption{Our submissions to the evaluation leaderboard, along with baseline results. The best results in each metric are underlined. Submission \#4 was our primary system.}
\begin{tabular}{llcccc}
\toprule
\# & Description                               & UAR$\uparrow$  (\%)  & F1$\uparrow$ (\%) & Accuracy$\uparrow$ (\%) & MAE$\downarrow$   \\ 
\midrule
& Baseline \#1 (SVM) \cite{getman2025non} & 22.1 &  N/A & 32.7 & 1.05 \\
& Baseline \#2 (MT w2v2) \cite{getman2025non} & 36.4 & N/A & 54.5 & 0.55 \\ \midrule
1  & GOP-CTC-AF, ordinal CE                       & 38.1      & 39.1    & 55.0          & \underline{0.505} \\
2  & Same as \#1, trained on full training set & 39.3      & 40.1    & 54.6          & 0.515 \\
3  & \#2 + E2E-R, interpolated                 & 36.6      & 37.4    & 52.4          & 0.591 \\
\textbf{4} & \textbf{GOP-CTC-AF-E2E, weighted ordinal CE}              & \underline{44.8}      & \underline{47.4}    & 55.8          & \underline{0.505} \\ 
5  & All 3 models, interpolated, optimized on dev                & 42.0      & 42.0    & \underline{56.4}          & 0.511 \\ \bottomrule
\end{tabular}
\end{table*}

The E2E-R training consists of the 2 stages. In the first stage we fine-tuned a wav2vec2.0-based end-to-end letter recognition model on the NB Tale dataset. The full model was optimized using a joint CTC-Attention loss with a CTC weight of 0.2. All components were trained using the Adam optimizer with a learning rate of 3e-4 for non-wav2vec2.0 parameters and 1e-4 for wav2vec2.0. Training was conducted for 6 epochs with  an effective batch size of 16. The learning rates were annealed using a NewBob scheduler with an improvement threshold of 25e-4. The best model was selected based on the phoneme error rate (PER) on the development set. In the second stage, training was conducted for 15 epochs with a batch size of 16. The learning rates were 2e-4 for the decoder and scoring modules, and  3e-6 for wav2vec2.0 parameters.  The training objective was unweighted ordinal CE loss. All model parameters were optimized using Adam optimizers with separate schedulers (NewBob) for the wav2vec2, decoder, and scoring components, triggered by validation MAE. The model’s performance was monitored on a development set, and the best checkpoint was selected based on the highest UAR.

For training the prefix-tuned and GOP-CTC-AF-E2E models, we applied speed perturbation to the training data using factors of $\pm10\%$. The models were first optimized on the internal training split of the official dataset and finally trained on the full speed-perturbed training set. Both models were trained using a learning rate of $10^{-3}$ and an effective batch size of 64. To stabilize training, a learning rate multiplier of 0.01 was applied to the pretrained wav2vec2.0 backbone. We used a learning rate warmup of 100 steps followed by a linear decay schedule. 
We used weighted ordinal CE loss, where class weights were set inversely proportional to class frequencies. A class distance scaling factor of $\alpha = 0.5$ was applied, effectively making the distance penalty proportional to the square root of the ordinal distance between the predicted and true labels. We found that using a more aggressive distance penalty (i.e., larger $\alpha$) led the model to avoid predicting extreme scores such as 1 or 5, likely due to the disproportionately high loss associated with large errors.

Training was conducted for 10 epochs, but the final model  was selected based on the best UAR on the internal development set. 
SpecAugment was used on the wav2vec2.0 features with the following parameters: feature masking block width of 64, block start probability of 0.4\%, time masking block width of 10, and time mask start probability of 6.5\%. 

\subsection{Results}

Table \ref{tab:dev-results} lists results of the three models on the internal development set.

Our five submissions to the evaluation leaderboard are listed in Table \ref{tab:eval-results}. The evaluation set predictions were generated using the following systems:
\begin{enumerate}[noitemsep,topsep=2pt,parsep=5pt,partopsep=2pt]
\item GOP-CTC-AF-E2E model, trained on the internal training split using ordinal CE loss with distance penalty scaler $\alpha = 1.5$, without class weights
\item Same as \#1, but trained on the full official training set. 
\item Predictons of \#2 interpolated with a E2E-R model, with interpolation weights (0.1, 0.9) optimized on the internal development set to maximize UAR.
\item GOP-CTC-AF-E2E model, trained using weighted ordinal CE loss ($\alpha=0.5$) on the full training set.
\item Interpolation of all three models (listed in Table \ref{tab:dev-results}), weights optimized on development data.
\end{enumerate}

Since submission \#4 resulted in the best UAR and MAE scores, we used this as our primary result in the leaderboard. On the internal development set, the three-model interpolation (\#5) outperformed model \#4  by a large margin, obtaining an UAR of 50.4\%. However, this didn't translate to better performance on test data, suggesting overfitting to the internal development set.

\subsection{Ablations}

We conducted a small ablation study to investigate the importance of different components in our best model (GOP-CTC-AF-E2E). Table \ref{tab:ablation} lists results on the internal development set when the following individual changes were made to the model: (1) freezing the pretrained wav2vec2.0-based CTC model, (2) removing the GOP Transformer from the model, (3) removing the canonical letter embeddings as additional features to the GOP Tranformer, (4) using simple CE loss instead of weighted ordinal CE loss. It can be seen that freezing the CTC model dramatically reduces model accuracy, while other ablations have smaller effect on the performance. Using CE loss results in high accuracy, but it reduces UAR and MAE performance.

\begin{table}[tb]
\centering
\label{tab:ablation}
\caption{Ablation results with the GOP-CTC-AF-E2E model.}
\begin{tabular}{l|ccc}
\toprule
                              & UAR  (\%) & Acc. (\%) & MAE$\downarrow$   \\ \midrule
GOP-CTC-AF-E2E    & \textbf{44.0}       &\textbf{54.7}          & \textbf{0.525}     \\ \midrule
- Frozen CTC model & 32.6      & 35.8 & 0.916 \\
- No GOP Transformer & 41.3 & 51.6 & 0.563 \\
- No char. embeddings & 42.2 & 52.5 & 0.550 \\
- Simple CE loss & 42.4 & 54.0 & 0.549 \\ \bottomrule
\end{tabular}
\end{table}

\section{Conclusion}

We investigated three distinct end-to-end approaches for automatic pronunciation assessment within the NOCASA 2025 Challenge framework. Our best-performing GOP-CTC-AF-E2E model successfully integrates alignment-free CTC-based GOP features with Transformer-based contextual modeling, achieving superior results compared to other architectures. The use of a weighted ordinal CE loss, explicitly accounting for ordinal prediction errors, further enhanced model performance, particularly in terms of the primary evaluation metric (UAR). Ablation analyses underscore the critical role of making the GOP-CTC-AF-E2E end-to-end trainable and confirm the effectiveness of different model design decisions.

\bibliographystyle{IEEEbib}
\bibliography{refs}

\end{document}